\documentclass[sn-mathphys-num]{sn-jnl}

\usepackage{graphicx}%
\usepackage{multirow}%
\usepackage{amsmath,amssymb,amsfonts}%
\usepackage{amsthm}%
\usepackage{mathrsfs}%
\usepackage[title]{appendix}%
\usepackage{xcolor}%
\usepackage{textcomp}%
\usepackage{manyfoot}%
\usepackage{booktabs}%
\usepackage{algorithm}%
\usepackage{algorithmicx}%
\usepackage{algpseudocode}%
\usepackage{listings}%

\theoremstyle{thmstyleone}%
\theoremstyle{thmstyletwo}%

\theoremstyle{thmstylethree}%

\begin{document}

\title[Article Title]{A Similarity-Based Oversampling Method for Multi-label Imbalanced Text Data}


\author*[1]{\fnm{Ismail Hakki} \sur{Karaman}}\email{ihkaraman0@gmail.com}

\author[2]{\fnm{Gulser} \sur{Koksal}}

\author[3]{\fnm{Levent} \sur{Eriskin}}

\author[1]{\fnm{Salih} \sur{Salihoglu}}

\affil[1]{\orgdiv{Department of Industrial Engineering}, \orgname{Middle East Technical University}, \orgaddress{\city{Ankara}, \country{Turkey}}}

\affil[2]{\orgdiv{Department of Industrial Engineering}, \orgname{TED University}, \orgaddress{\city{Ankara}, \country{Turkey}}}

\affil[3]{\orgdiv{Department of Industrial Engineering}, \orgname{Piri Reis University}, \orgaddress{\city{Istanbul}, \country{Turkey}}}


\abstract{In real-world applications, as data availability increases, obtaining labeled data for machine learning (ML) projects remains challenging due to the high costs and intensive efforts required for data annotation. Many ML projects, particularly those focused on multi-label classification, also grapple with data imbalance issues, where certain classes may lack sufficient data to train effective classifiers. This study introduces and examines a novel oversampling method for multi-label text classification, designed to address performance challenges associated with data imbalance. The proposed method identifies potential new samples from unlabeled data by leveraging similarity measures between instances. By iteratively searching the unlabeled dataset, the method locates instances similar to those in underrepresented classes and evaluates their contribution to classifier performance enhancement. Instances that demonstrate performance improvement are then added to the labeled dataset. Experimental results indicate that the proposed approach effectively enhances classifier performance post-oversampling.}

\keywords{oversampling, text classification, multi-label classification, imbalanced classification, text similarity}



\maketitle

\section{Introduction}\label{intro}

Unstructured data, such as images, texts, videos, sensor readings, tweets, logs, audio, and emails, forms a significant portion of the data generated today. Text classification remains one of the most widely used tasks in Natural Language Processing (NLP). \cite{aggarwal2012survey} remarks that assigning a given text such as sentences, paragraphs, and documents to predefined categories with the help of ML has countless advantages in different domains. Classifying documents into topics, classifying emails for the related departments, and analyzing positive and negative comments for a product are some examples of application areas of text classification that provide great convenience by reducing human effort and thus eliminating human errors. Although it has many different applications and the number of studies and projects increases day by day, there are also some challenges in text classification. Firstly, the structure of text data is complex, as it consists of semantic elements that require sophisticated techniques for computer interpretation. Secondly, acquiring labeled data, especially in specialized domains like law and medicine, is challenging. Labeling text data demands considerable cognitive effort and domain expertise. Therefore, solutions that streamline the labeling process would be highly beneficial.

In some classification tasks, data can have multiple labels simultaneously. For instance, a news article might be categorized under both the economy and foreign relations. Similarly, in an image labeling project featuring animals, a single picture may contain more than one animal. When instances belong to multiple classes, the task is referred to as multi-label classification. In this scenario, labels are represented as multi-dimensional vectors rather than single classes or categories. In real-world applications, some classes may have a large number of instances, while others may have insufficient data for training a classifier. When classification algorithms are trained on imbalanced or limited datasets, achieving satisfactory results for underrepresented classes becomes difficult. This issue is particularly pronounced for complex machine learning models, which often require substantial amounts of data to perform effectively. For text classification tasks, the complexity of the data typically necessitates the use of advanced models, underscoring the critical need for adequate data to create effective classification systems. This challenge, known as the data imbalance problem, is one of the most significant hurdles in classification tasks.

\begin{figure}[h]
\center
\includegraphics[scale=0.55]{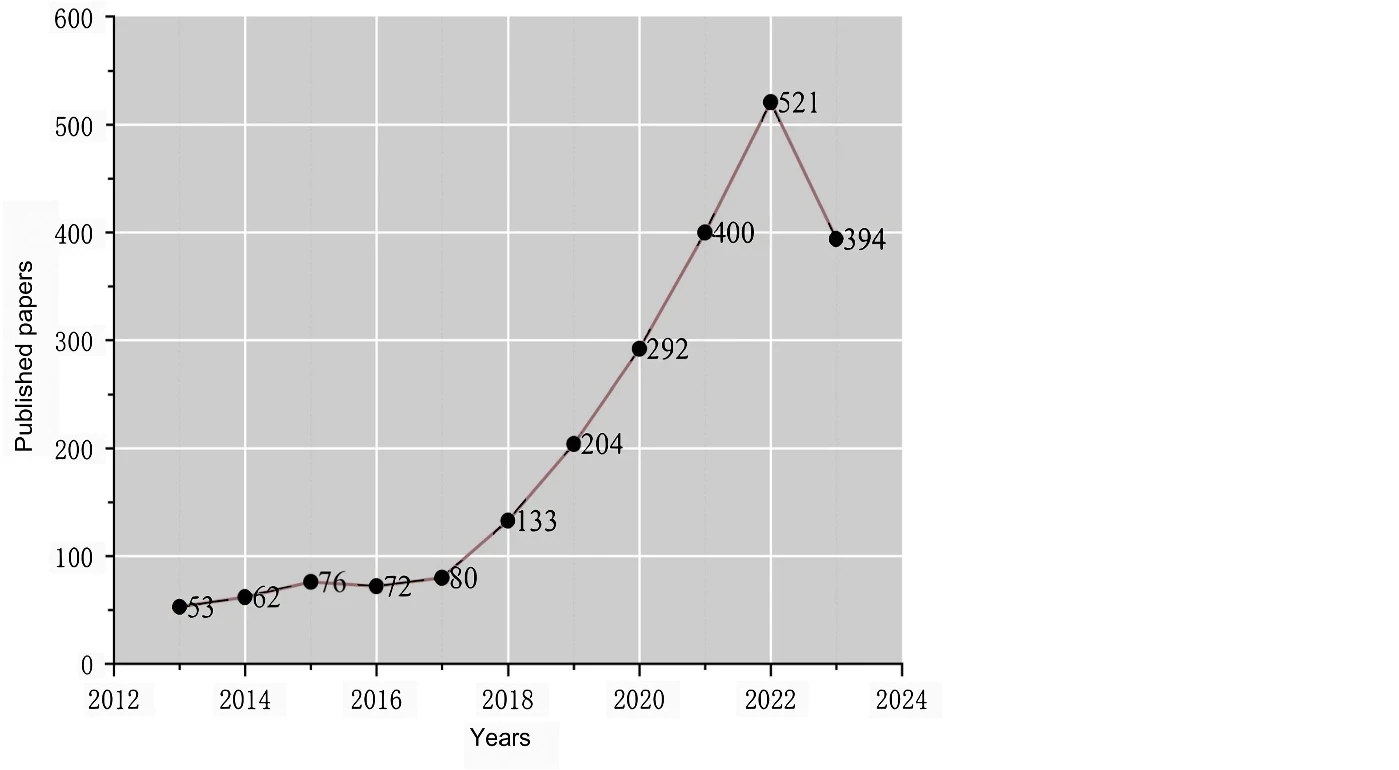}
\caption{The number of publications in imbalanced learning (Reprinted from \cite{chen2024survey}). }
\label{stud_dist}
\end{figure}

The class imbalance problem has received considerable attention in machine learning and pattern recognition due to the problems it poses \cite{mollineda2007class}. It is more important in such real-world applications where misclassifying a minority example costs a lot. Spam mail classification, diagnosis of rare diseases, and authentic document classification are the main examples where the cost of classifying rare instances is very high. Different solutions have been proposed in the literature to avoid the effect of imbalanced data in such critical problems. The trend for the published papers in the literature is presented in Figure \ref{stud_dist}. We can see that the number of papers has increased greatly in recent years. 

A primary challenge in achieving balanced datasets in real-world applications is the inherent difficulty of obtaining such data. This challenge is particularly pronounced in multi-label classification tasks, where class imbalance issues are often severe. A straightforward approach to addressing this issue is to acquire additional labeled data; however, restarting the labeling process is often unfeasible due to the associated high costs and resource demands \cite{fredriksson2020data}. In fact, data science projects allocate approximately 80\% of their efforts to data collection, labeling, and preparation. This intensive process is unavoidable as most datasets are initially unlabeled, inadequately labeled, or insufficiently representative to train an effective model. The complexity of labeling tasks is further compounded when dealing with textual data, one of the most challenging data types to annotate. Unlike images, videos, or numerical data, text data requires a nuanced understanding of semantics, demanding time-consuming reading and comprehension. Additionally, to ensure reliable annotations, domain expertise is often essential. Annotators must possess in-depth knowledge of the relevant field. For instance, legal document annotation necessitates input from legal experts, as the language and concepts in legal texts are complex and inaccessible to non-specialists.

Moreover, even if additional data is collected, it is unlikely to resolve issues of data sufficiency and imbalance. Real-world data distributions tend to mirror those found in existing datasets, meaning that new data would typically have similar class distributions. For example, in a dataset where only five out of a thousand labels represent a minority class, adding thousands of additional instances may still fail to achieve a balanced representation for that class. This scenario underscores that, despite substantial investments in time and resources, the label distribution often remains unchanged, and the problem of insufficient and imbalanced data persists.
    
As an alternative to solve the data imbalance problem, the so-popular self-supervised methods are proposed to utilize unlabeled data at hand. In self-supervised methods, a classifier is trained with a small amount of labeled data and uses the trained classifier to predict unlabeled samples and label new samples. Thus, the labeled dataset is extended using the trained model. In the literature, some studies focused on this specific area and proposed several methods. However, the main drawback of these methods is relying on an initial classifier that might have a poor performance. If the performance of the model is satisfactory, then there is no need to improve the dataset. Conversely, if the initial results are not satisfactory, then it is not possible to trust this poorly performing classifier.  

The pervasive issue of class imbalance in real-world datasets, particularly in multi-label settings, underscores the need for a robust and comprehensive solution. In multi-label classification, the imbalance problem is further complicated by the inherent nature of the task, where multiple labels can be assigned to a single instance. This overlap of classes precludes clear-cut boundaries between them, which poses significant challenges for traditional oversampling methods. Classical oversampling approaches typically focus on balancing classes individually by generating additional instances within each class. However, this approach is ineffective in multi-label classification settings, where distinct separation between classes does not exist. When an instance belongs to multiple classes simultaneously, using it as a reference in the oversampling process introduces ambiguity, as it is unclear whether the newly generated instances should belong to all associated classes or only a subset. Consequently, the lack of clear class boundaries in multi-label data prevents conventional methods from effectively addressing the imbalance issue, highlighting the need for innovative approaches tailored to the unique challenges of multi-label classification. 

To solve this problem in multi-label imbalanced text data classification tasks, an auto-labeling algorithm that labels the unlabeled instances by utilizing the similarity between the instances is proposed. The proposed algorithm finds instances similar to the current ones in the data space and considers them as candidate instances. If candidate instances help improve the overall performance, they are added to the labeled set. The algorithm searches the unlabeled data space iteratively and finds the possible instances to extend the labeled set.  

The paper is organized as follows: in Section \ref{chp:b2}, we provide a comprehensive review of the literature and relevant background. Section \ref{proposedmethod} outlines the proposed similarity-based oversampling method, including its motivations and methodological framework. Section \ref{chp:b4} presents the experimental setup, detailing the datasets, evaluation metrics, and performance analyses. Finally, in Section \ref{conclusion}, we discuss the findings, implications, and future research directions based on the results of our study. 

\section{Literature Review and Background}
\label{chp:b2}

\subsection{Multi-label Text Classification}

\begin{figure}[H]
	\center
    \includegraphics[width=0.8\textwidth]{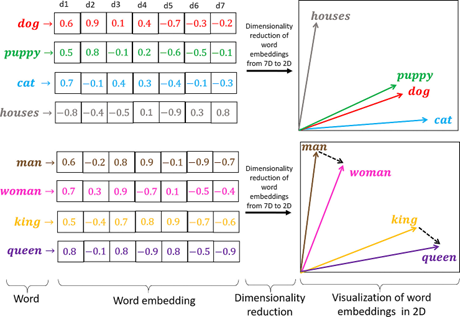}
    \caption{Embedding representation for some words.}
    \label{embed_rep}
\end{figure}

ML algorithms rely on numeric data, making the conversion of text to numeric is a key challenge in text classification. Techniques for this conversion range from simple methods like one-hot encoding to complex models such as transformers including Bag of Words, skip-gram, word2vec, and tf-idf vectorization \cite{singh2019vectorization}. Simple methods often lead to sparse representations, causing efficiency issues and inadequate semantic representation. To address these, word embeddings convert words into dense vectors using complex neural networks, with the powerful ability to capture semantic relationships, definitions, and context. The variety of embedding methods is expanding with proven success in NLP tasks. Notable examples include Word2Vec, BERT, GloVE, fastText, ELMo, GPT, and NPLM \cite{asudani2023impact}. A representation of word embeddings is shown in Figure \ref{embed_rep}.
 
Embeddings are a powerful representation method for text data and are applicable across various types of problems \cite{ruzzetti2021lacking, salihoglu2024enhancing}. They capture the semantics of words and the relationships between them. To apply embedding techniques, text data must undergo preprocessing to clean and standardize it \cite{anandarajan_hill_nolan_1970}. Key steps include removing meaningless elements like numbers, emails, stop words, and punctuation. The specific operations may vary based on the content and type of data. For example, web-scraped data often contains noise like HTML tags and extraneous characters. According to \cite{kowsari2019text}, once text is standardized, tokenization is applied to break the text into smaller units like sentences or words. The next step in text classification is identifying root words \cite{forman2003extensive}. Stemming reduces words to their root form, allowing different variations—like “working,” “worker,” and “works”—to be stemmed to “work.” However, stemming can lead to loss of meaning, as in the case of “working” and “worker.” An alternative is lemmatization, which converts words to their base form, preserving their meanings \cite{balakrishnan2014stemming}. Choosing one of these techniques prepares the text for numerical conversion and subsequent feeding into ML models.

It is still a major research topic and in the literature, almost any kind of classifier is tried to examine results to find the best algorithm. Naive Bayes is one of the most popular algorithms chosen in text classification tasks \cite{kim2002effective}. Due to its mechanism that captures the probabilistic relations, it tries to find the pattern between word occurrences and output. However, while the complexity of embeddings increases Naive Bayes loses its ability to capture that pattern. Support Vector Machines (SVMs) work well with high-dimensional datasets so it is widely preferred for text classification tasks \cite{chau2008machine,joachims1998text,wang2006optimal,goudjil2018novel}. SVM tries to find a decision boundary in the data space if it exists. Its main drawback is that it requires computation power and is slow for training compared to other methods \cite{yuan2008msvm}. On the other hand, the linear version of SVM performs well and it is very efficient to train compared to standard SVM \cite{dumais1998using}. Decision Tree is another classification algorithm widely used in different areas and also in text classification tasks \cite{lewis1994comparison}. Its tree-based structure helps to create decision rules to decompose the data space in a very fast-paced and easy-to-interpret algorithm. On the other hand, it tends to overfit the data easily. Random Forest is an ensemble algorithm composed of many decision trees in parallel \cite{ho1995random}. Predictions from parallel decision trees are combined to create a final prediction via the voting scheme. Random Forest performs well in most of the tasks, as well as text classification. Since it is an ensemble algorithm, its computation time is high compared to other algorithms. Logistic Regression is the modified version of linear regression by adding a sigmoid function to the output layer. The predicted real values are converted to classes by using the sigmoid function. Logistic regression is also widely used in text classification problems \cite{pranckevivcius2016application}. 

Today, in most of tasks, Deep Learning (DL) achieves state-of-the-art results, driven by advancements in computational power and data availability. Since it is an emerging study field, several studies in the literature study different architectures to improve the performance of various NLP tasks. With the help of a lot of data, large language models have versatile applications beyond conversational tasks such as coding, data generation, and other complex computational tasks \cite{gurdil2024integrating, kalla2023study}. On the other hand, when resources are limited, it can be difficult to train a model. \cite{liu2024knowledge} builds on recent advancements in knowledge-enhanced prompt learning methods for few-shot text classification, which aim to improve classifier performance in low-resource settings by integrating structured knowledge into prompt templates and verbalizers. Key architectures used in the NLP world include Recurrent Neural Networks, Long Short-Term Memory Networks, Convolutional Neural Networks, and Transformers \cite{gupta2020compression, singh2021nlp}. However, DL models require a huge amount of data for training. So, when the main problem is data insufficiency, it is not a good choice to use a DL model at all.  

Multi-label classification is a type of classification problem when the instances belong to more than one class. Since the output is different from other classification methods, the traditional evaluation measures do not work for multi-label settings. They should be modified or new measures should be introduced to evaluate the performance. In the literature, modified versions of existing measures and also, new measures designed for multi-label classification are used \cite{tsoumakas2007multi, ganda2018survey, zhang2013review}. The modified versions of traditional measures, and new measures specially designed for multi-label settings are introduced in these studies. Overall, measures can be categorized under classification measures and ranking measures. Classification measures are interested in the performance of the predictions. Ranking-based measures are designed to measure the performance of the prediction probabilities. 

The exact match ratio (EMR), also known as subset accuracy, is the proportion of correctly predicted label set, i.e. label vector, to the total number of label vectors. It considers the label set as a whole and counts only if all the labels of an instance are predicted correctly, which makes it a very strict measure \cite{ganda2018survey}. Accuracy measures the ratio of correctly predicted individual labels to the total number of labels \cite{nazmi2020evolving}. Compared to subset accuracy, it considers all the components of the label vector and a more indulgent version of it. Hamming loss is the proportion of individual labels that are incorrectly predicted \cite{ganda2018survey}. It simply equals one minus the accuracy score.
    
Precision is found by dividing the number of correctly predicted outputs by the total number of predicted outputs. The proportion of correctly predicted outputs to the total number of true outputs is called recall \cite{gibaja2015tutorial}. There is a trade-off between precision and recall. If the focus is on precision, an increase in precision will cause a decrease in the recall and vice versa. On the other hand, if the desire is to obtain accurate predictions, then the precision should be maximized. F1-score solves this trade-off and focuses on precision and recall at the same time. As it can be seen from Equation \ref{eqn:F1score}, it is the harmonic mean of precision and recall. If there is no special focus on precision or recall, F1-score is generally a better measure in classification tasks.

    \begin{equation}
    F1\mbox{-}score = \frac{2*Precision*Recall}{Precision+Recall}
    \label{eqn:F1score}    
    \end{equation}

One error is the ratio of examples whose top-ranked label is not in the true label set \cite{tarekegn2021review}. It only cares about the top-ranked label. It is a loss measure and should be minimized. Coverage computes how far it is needed to go through the ranked scores to cover all true labels. The rank of a label is calculated by ranking the prediction probabilities and finding the number of labels whose probability is greater than or equal to its score. The best value is equal to the average number of labels in the true label set per sample \cite{tsoumakas2009mining}. Ranking loss is the average number of label pairs that are incorrectly ordered given that the prediction probabilities are weighted by the size of the label set and the number of labels, not in the label set \cite{gao2011consistency}. It is similar to error set size with a difference in weighting with relevant and irrelevant labels. Since it measures the loss, the best value is zero. Lastly, the average precision is the weighted mean of precisions achieved at each threshold level on a precision-recall curve. The weight is the increase in recall from the previous threshold used as the weight \cite{liu2015multi}. 
    
Another aspect of our problem is the imbalancely distributed classes. The imbalanced class distribution causes a deviation between the class’ performances and the weighting methods are needed to incorporate the importance of the classes \cite{zhang2013review}. Micro averaging calculates the measures globally, without separating classes. On the other hand, macro averaging calculates the measure for each class and averages them as they have equal weights. Macro averaging works well when all the classes have equal weights. Also, weighted averaging is a method that calculates measures for each class and takes the weighted average, proportional to the number of instances in a class. Lastly, sample averaging calculates the measures for each instance and finds their average. It does not discriminate against classes and considers all the instances equally important. 
    
Quantifying the performance of models correctly has vital importance in case of data imbalance. The performance measure should meet the requirements of the problem. Choosing the wrong measure will mislead the evaluations and result in choosing a poor model. Generally, precision, recall, or F1-score are preferred over accuracy for imbalanced classification problems \cite{feng2021imbalanced}.

\subsection{Imbalanced Data Classification}
    
In \cite{kaur2019systematic}, a comprehensive review of 152 articles is made to organize the proposed solutions for the imbalanced data classification problem. The main approach types of imbalanced learning are preprocessing methods, cost-sensitive learning methods, algorithm-centered approaches, and hybrid methods. Also, \cite{haixiang2017learning} analyzes 527 papers and they found that nearly 30\% of the papers are published in resampling techniques which proves resampling is the most popular oversampling technique. Resampling techniques are for generating new instances by utilizing the existing ones. While undersampling removes the instances from the majority class, oversampling generates new synthetic instances with the help of existing instances. Hybrid methods employ both strategies at the same time to yield better results.

A well-known example of an oversampling method is the Synthetic Minority Oversampling Technique (SMOTE) \cite{chawla2002smote}. SMOTE creates new synthetic instances by the linear combination of current instances in the same neighborhood. Also, different versions of SMOTE are proposed like Borderline-SMOTE, DBSMOTE, MWMOTE, etc. \cite{fernandez2018smote}. Unfortunately, there are some drawbacks to SMOTE in multi-label settings. When there is overlapping between classes, it is hard to find a distinction between different classes. Two different LP transformation-based resampling approaches are proposed in \cite{charte2013first}, LP-RUS for undersampling and LP-ROS for oversampling. LP-ROS is the proposed oversampling method that clones the instances in minority label sets. These methods do not work well when the label combinations are distinct. \cite{charte2015addressing} proposes one undersampling and one oversampling method that focuses on the frequency of individual labels instead of label sets are proposed. ML-RUS deletes the instances in the majority class and ML-ROS clones samples in the minority class. On the other hand, these methods do not work well when the joint occurrence of minority and majority labels exists. In \cite{charte2019dealing}, REMEDIAL is proposed to overcome this problem by decoupling majority and minority classes. There are also heuristic approaches, such as MLeNN, proposed in \cite{charte2014mlenn}, which utilizes the Nearest Neighbor algorithm, and MLTL \cite{pereira2020mltl}, which employs the Tomek Link algorithm, among others. The problem with oversampling algorithms, they generally copy the existing instances, and copied instances don’t add any predictive power to the algorithm so the lack of data problem is still unsolved. 
    
In \cite{luo2019novel}, a novel oversampling method for text data using sequential generative adversarial networks (seqGAN) is proposed. The GAN model consists of a generator that creates data to confuse the discriminator and a discriminator that distinguishes between generated and original data. This competitive process improves the generator’s performance, and experiments show that it outperforms traditional oversampling methods such as random oversampling and SMOTE. Additionally, \cite{shaikh2021towards} introduces a text generation algorithm combining GPT-2 and LSTM. This method produces a new text that maintains grammatical structure and semantic coherence. LSTM excels with short texts, while GPT-2 performs well with longer documents. However, both approaches are based on complex DL models that require substantial initial data. In \cite{chen2014sentence}, a continuous skip-gram model is trained using word2vec to obtain word/POS pairs, addressing the different lexical functions of words based on their POS. Sentence vectors are created by summing word vectors, and new sentence vectors are formed by combining pairs from minority classes. However, the assumption that summing two vectors yields a valid class vector may not hold true. Moreover, \cite{moreo2016distributional} suggests a version of latent space oversampling; distributional random oversampling is proposed. A generative function is created based on the distributional properties of the documents to return a vectorial representation of words to create new samples. \cite{mohasseb2018improving} uses SMOTE to oversample the question data to balance the data ratio. NB classification is used to classify instances into categories.

Most of the proposed methods work with numerical data by converting text data into numerical format. Some methods utilize raw text data by modifying it with synonyms or adding/removing random words. The TextAttack framework \cite{morris2020textattack} supports adversarial attacks, data augmentation, and adversarial training for text. Its data augmentation tool generates perturbed versions of input text while ensuring their validity, effectively expanding the training dataset. Similarly, EasyAug \cite{qiu2020easyaug} helps users compare various text data augmentation methods, employing techniques like random oversampling, word-level transformations, and variational encoding. Additionally, \cite{li2018imbalanced} proposes an oversampling method for minority class texts using two strategies: inversion and imitation. In the inversion process, samples from the majority class are modified by reversing sentimental words, while imitation creates new samples by replacing words in minority-class texts with semantically similar alternatives. Although synthetic texts may lack grammatical correctness, they should retain some semantic interpretability.

Additionally, \cite{jang2021sequential} introduces an adaptive distribution selection strategy that overcomes the limitations of under-sampling and over-sampling using continuous learning. This method partitions the training dataset into exclusive subsets, enabling sequential learning aligned with the target distribution while retaining knowledge from previous tasks. \cite{shi2020penalized} presents a Multiple Distribution Selection (MDS) approach, which combines a single softmax distribution with degenerate distributions to better model complex and unbalanced data. Words are first converted into pre-trained word embeddings, and then max-pooling is used to construct sentence embeddings. MDS automatically learns distribution weights in the second stage, outperforming traditional cost-sensitive methods. \cite{tian2020graph} employs graph models to address imbalance issues. In this method, the data set is transformed into a graph model, calculating overlapping and disjunct degrees to derive a graph-based imbalance index. This index analyzes the characteristics of intrinsic data to improve performance in unbalanced classification.

\cite{pavlinek2017text} proposed a semi-supervised self-training method based on LDA for document classification. The ST LDA algorithm uses a small labeled set and a larger unlabeled set to train a classifier. It expands the labeled set by adding similar instances from the unlabeled data and involves two key steps: self-training to enlarge the dataset and parameter tuning. In \cite{wu2017highly}, a differential-evolution semi-supervised classification algorithm is introduced. It starts by initializing the model's parameters and training classifiers using the labeled dataset. High-confidence instances from the unlabeled set are predicted and combined with the labeled set. A differential-evolution-based selection then optimizes the predicted instances, extending the original labeled set.

In \cite{meng2018weakly}, a weak-supervised method is proposed. It generates new documents using the labeled ones and trains a model to share its parameters with a self-training classifier to label unlabeled documents. \cite{guzman2009using} proposes a new self-training method for text classification that utilizes automatic extraction of unlabeled text from the web. After acquiring new data instances from the web a semi-supervised method is utilized to assign new instances to classes. Also, \cite{zhang2007semi, karisani2021semi, lanquillon2000partially, xu2021semantic} are the other studies that employ self-training for text classification. 

Also, there are some alternative approaches proposed for automated data labeling. \cite{desmond2021semi} uses the active learning approach to label unlabeled data. Firstly, K-Means clustering is used to partition data into clusters. Data is iteratively served to the labeler and label spreading is applied to calculate probabilities for the rest of the data. \cite{zhang2021lancet, hajmohammadi2015combination} also employ active learning to automate the labeling process with the help of an annotator. \cite{luo2019deep, li2021combining} are the methods that combine self-learning with DL. In \cite{luo2019deep}, a multi-labeled medical patent classification approach is studied. After preprocessing text, using Glove, embeddings are obtained and architecture is developed containing a bi-directional LSTM and a fully connected layer. Another method that is used in this context is zero-shot learning \cite{meng2020text, ye2020zero}. In zero-shot learning, a model is trained on a dataset with specific labels, and this method is used to predict instances that belong to unseen classes. Furthermore, some studies combine different approaches with self-training such as K-NN, evolutionary algorithms, and few-shot learning \cite{li2020boosting, donyavi2020using, mukherjee2020uncertainty}.

There are some drawbacks to self-supervised methods that make them impracticable in some cases. First, if the data is not enough to train an initial classifier with acceptable results, it is not possible to make predictions for labeling new instances. Second, self-labeled techniques follow an iterative procedure, aiming to obtain an enlarged labeled data set, in which they accept their predictions to be correct. If the trained classifier does not perform well on the initial set then the predicted labels will not be accurate as well. The second case is also related to the first one because when there is not enough data to train a classifier, that classifier performs poorly. As mentioned previously, our starting point is that the labeled data at hand is so insufficient that it is not possible to train an acceptable initial classifier. When it is not possible to train an initial classifier, self-supervised methods become impractical. Especially, the possibility of this problem in the multi-label case is quite high. 
    
\subsection{Similarity Functions for Text Data}

Similarity measures are functions that are used to measure the matching score between two objects. The more similar a pair of objects, the higher the similarity score produced by the similarity measure and can be considered as the inverse of the distance function. In NLP, similarity measures are used to measure the similarity between documents, sentences, or words. It plays a crucial role in designing our algorithm, as it helps identify new instances. Choosing the appropriate similarity function is essential; it must accurately reflect the relationships between embedding vectors that represent text data in a high-dimensional vector space, capturing its semantics. The similarity between these embedding vectors indicates the semantic similarity of the text, allowing us to identify similar instances that belong to the same class.

The similarity of texts can be measured in two ways \cite{wang2020measurement}. First, lexical similarity analyzes the resemblance of the words and their sequence, and, ‘apple’ and ‘apply’ are considered as lexically similar. On the other hand, semantic similarity measures the similarity between meanings using knowledge-based or corpus-based models are used to measure semantic similarity. ‘apple’ and ‘apply’ are not semantically similar because they represent completely different things.

The most popular way to measure semantic similarity is length distances. Euclidean, cosine, Minkowski family, Jensen Shannon distance, and Hamming distances are the widely used length distance types \cite{deza2009encyclopedia,norouzi2012hamming}. In \cite{magara2018comparative}, they studied text similarity measures and conducted a comparative analysis between them. They measure the similarity between research papers for a recommendation algorithm using different similarity functions and according to their experiments, cosine similarity yields the best results. 

\section{Proposed Method}
\label{proposedmethod}

\subsection{Motivation}
\label{motivation}

In the previous chapter, we examined the need for auto-labeling methods to mitigate the challenges of imbalanced data with minimal manual effort. Real-world datasets often exhibit significant class imbalance and are insufficient for training machine learning models, which can lead to suboptimal performance, particularly in text classification tasks. Traditional approaches, such as manually labeling additional data, are often impractical due to the complexity of domain-specific requirements and the need for expert annotators. To address these issues, we propose a similarity-based oversampling method that identifies new instances from an unlabeled dataset to address data imbalance and insufficiency in multi-label text classification. This method iteratively searches the unlabeled dataset for similar instances, using a similarity measure based on word embeddings. By applying a class similarity threshold, the method identifies candidate labels for these instances and then evaluates whether incorporating them enhances the model’s performance. Instances that contribute to performance improvements are then added to the labeled set. This approach represents a novel contribution to the literature, filling a gap in current methodologies and offering a foundational technique that future algorithms can build upon.

\subsection{The Overall Algorithm}
\label{overall_alg}

In our proposed method, as in the self-training approaches, we aim to oversample the dataset by using unlabeled instances by utilizing a similarity function between instances. In self-training approaches, an initial classifier is trained to predict unlabeled instance labels to add them to the labeled set. However, we cannot rely on the labels predicted by the initial classifier for some classes due to the lack of enough labeled instances for those classes in the training set. So, after converting text data to vector embeddings, by utilizing a similarity function that maximizes the within-class similarity while minimizing the between-class similarity, it is possible to find new instances that have high similarity to the labeled set. The main motivation for using the similarity function is to ensure that instances belonging to the same class are positioned close to each other in the vector space, while instances from different classes are relatively farther apart.

Our proposed solution is designed for the case when the below issues arise:
\begin{enumerate}
  \item When the labeled data at hand is insufficient to train a classifier and/or the result is poor.
  \item Labeling more data is not possible and/or also, labeling more data will not solve the data imbalance problem because the real distribution of the data is also highly imbalanced as discussed in Chapter \ref{chp:b2}.
  \item Labeling data requires more effort/expertise than collecting.
  \item The data is in a multi-labeled format.
  \item The relation between instances can be captured by a similarity function.
\end{enumerate}
 
To begin, the dataset is split into training and test sets before any preprocessing to prevent data leakage. The training set is used for preprocessing, oversampling with our proposed algorithm, and training the classifier, while the test set, which remains unseen during training, is reserved exclusively for evaluation purposes. Following the dataset split, a preprocessing step is applied to clean, standardize, and structure the text before converting it into embeddings. This cleaning process involves converting text to lowercase, removing punctuation, HTML tags, URLs, stop words, and tokenization. Next, the text data is converted into embeddings using popular techniques, including TF-IDF, Word2Vec, GloVe, Hugging Face models, ELMo, BERT, and ChatGPT Embeddings. We experimented with various embedding methods, which are detailed in Appendix \ref{app1}. However, as evaluating different embedding techniques lies outside the scope of this study, we ultimately selected the 'all-roberta-large-v1' model for its strong overall performance. The primary performance criterion in this context is the optimization potential of similarity measures between instances.

\begin{figure}[h]
	\center
    \includegraphics[width=0.8\textwidth]{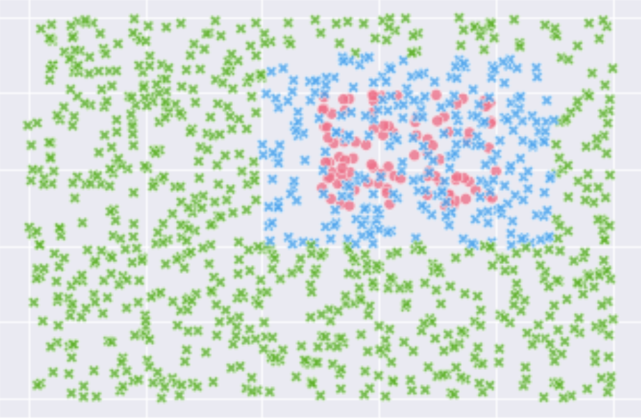}
    \caption{A representation of the labeled and unlabeled instances.}
    \label{repr}
\end{figure}

To enhance the performance of the oversampling method, it is essential to accurately distinguish within-class and between-class similarity. Ideally, instances belonging to the same class should exhibit high similarity scores, while those in different classes should have low similarity scores. Word embeddings convert text into vectors such that similar words are positioned near each other within the vector space. A crucial aspect of our approach is leveraging this property of word embeddings to identify similar instances from the unlabeled set using a similarity function. Instances within the same class, or those with similar semantic meanings, naturally tend to cluster in the vector space \cite{clark2015vector}. Furthermore, fine-tuning the embeddings can reinforce this clustering, ensuring that vectors of instances from the same class are closer together than those from different classes. This refinement makes within-class and between-class similarities more distinguishable, facilitating the accurate identification of instances belonging to different classes. Figure \ref{repr} represents a two-dimensional vector space to illustrate this phenomenon. The instances that are represented with red circles belong to the same class, hence, they are located close to each other. The unlabeled instances, which are marked as a cross, are distributed all around the space but our algorithm will try to find the closer instances from the potential unlabeled instances that are marked with blue.

\begin{figure}[H]
	\center
    \includegraphics[width=0.9\textwidth]{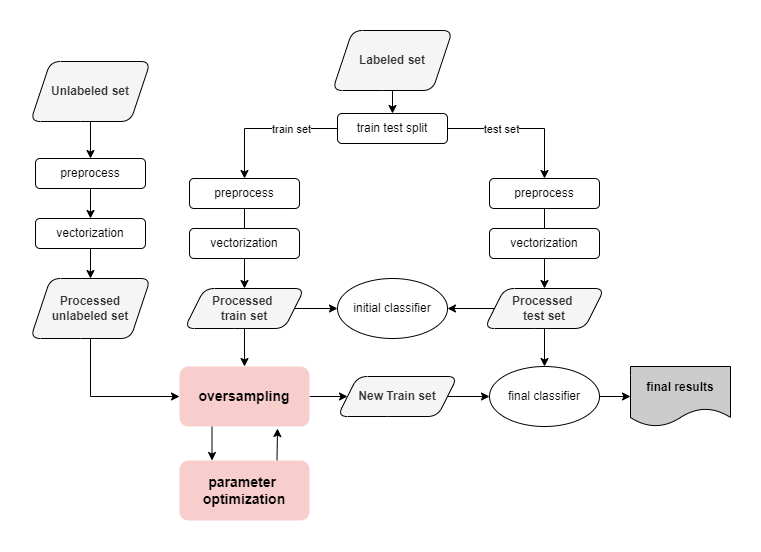}
    \caption{Flow chart for the overall algorithm.}
    \label{overs}
\end{figure}

The next step is optimizing the parameters of the oversampling algorithm. It has several parameters and the performance of the oversampling algorithm strongly depends on the parameter setting. It is not possible to make a suggestion for parameters since different datasets have different characteristics and the parameter settings can vary for each dataset. So, we strongly suggest optimizing the parameters to improve the performance of the algorithm. Each parameter of the algorithm and possible values they can take is explained in Section \ref{subsec:oversamp_alg}.

As a reference point, to evaluate the performance of the proposed methodology in terms of determining new instances efficiently, a control mechanism is required. To measure the performance of the oversampling algorithm, an initial and a final classifier are trained to observe the performance. If the performance of the final classifier improves with respect to that of the initial classifier, we can claim that the proposed methodology works efficiently. The flow chart of the overall algorithm is presented in Figure \ref{overs}. The pseudocode of the overall algorithm can be seen in Algorithm \ref{alg:overall_alg}.

\begin{algorithm}
    \caption{The Overall Algorithm}
    \label{alg:overall_alg}
    \begin{algorithmic}[1]
    \State Split labeled set as train and test
    \State Apply text preprocessing and convert to numeric form (unlabeled, train and test sets)
        \State \hspace{\algorithmicindent} Lowering and noise removal
        \State \hspace{\algorithmicindent} Stop words removal
        \State \hspace{\algorithmicindent} Tokenization and lemmatization
        \State \hspace{\algorithmicindent} Vectorization
    \State Fine-tune embeddings for training set to optimize within-class and between-class similarities
    \State Train an initial classifier as a reference point
    \State Parameter optimization and oversampling
            \State \hspace{\algorithmicindent} Design an experiment to optimize parameters
            \State \hspace{\algorithmicindent} Oversampling the dataset using the Algorithm \ref{alg:oversampling_alg} with optimized parameters
    \State Train a final classifier to compare results
    \State Yield the new labeled set
    \end{algorithmic}
\end{algorithm}

\subsection{The Oversampling Algorithm}
\label{subsec:oversamp_alg}

The proposed oversampling algorithm given in Algorithm \ref{alg:oversampling_alg} aims at finding new instances from the unlabeled vector set by utilizing the similarity scores calculated with the labeled vector set. It searches the unlabeled set and finds instances that are close to the existing ones as depicted in Figure \ref{repr}. These closely placed instances are candidate instances for labeling to extend the labeled set. An improvement checking procedure is used to check whether newly added candidates improve the overall performance or not. If the performance improves, candidates are added to the labeled set. Iteratively, the unlabeled set is searched to find new instances until several iterations are repeated. The pseudocode for our oversampling algorithm is presented below.

\begin{algorithm}
    \caption{The Oversampling Algorithm}
    \label{alg:oversampling_alg}
    \begin{algorithmic}[1]

        \State Calculate class similarities
        \State Calculate similarity factors 
        \State Train a baseline classifier
            \For{\text{number of iterations}}
                \State Calculate the required number of instances with a performance measure
                \State Find candidate instances
                \State \hspace{\algorithmicindent} Calculate selection probabilities for classes using the required number of instances
                \State \hspace{\algorithmicindent} Select a class randomly using the selection probabilities
                \State \hspace{\algorithmicindent} Find a batch of instances from the unlabeled set that has greater similarity than the similarity threshold calculated by class similarity times similarity factor
                \State Train a new classifier to look for improvement
                \State \hspace{\algorithmicindent} If the results get better 
                \State \hspace{\algorithmicindent} \hspace{\algorithmicindent} Add instances to the labeled set
                \State \hspace{\algorithmicindent} \hspace{\algorithmicindent} Increase similarity factors
                \State \hspace{\algorithmicindent} If the results do not get better
                \State \hspace{\algorithmicindent} \hspace{\algorithmicindent} Decrease similarity factors
                \State Update unlabeled set and iteration number
            \EndFor
    \end{algorithmic}
\end{algorithm}

The oversampling algorithm has some parameters that shape the functioning of the algorithm. Similarity calculation type, similarity function, batch size, number of iterations, balance ratio, and performance measure are the parameters of the algorithm. The algorithm takes the labeled set, unlabeled set, and the parameters to oversample the dataset and it returns the new labeled set, new unlabeled set, and metric history as output. The new labeled set is an oversampled set and the unlabeled set is the remaining unlabeled set that the algorithm did not consider as similar points. Metric history is an array and it is the list of performance measures that are measured during the oversampling step at each iteration.

The preparation step of the algorithm starts with calculating the class similarities between the word embeddings by using the binary combination of all instances in a class. The similarity function is a parameter of our method and we preferred to use cosine and Euclidean to keep it simple as they are the most popular ones. After calculating the similarity between vector pairs, an array of similarity scores is present at hand and there are two alternatives to degrade this array of similarities to a single score to represent class similarity. This is another parameter of the algorithm which is called similarity calculation type. It has two alternatives: ‘average’ and ‘safe interval’. If ‘average’ is used, the average of the array is calculated. Alternatively, ‘safe interval’ is tighter compared to ‘average’ and it finds the 3rd quartile of the similarity score array. The ‘safe interval’ will produce higher scores compared to the ‘average’ because the mean is expected to be less than the 3rd quartile in the normal distribution.

Afterward, similarity factors are computed based on class similarities, serving as an adjustable parameter for determining whether an instance is a viable candidate for labeling. Class similarities are fixed values, calculated during the training phase, and remain constant throughout the iterations. The similarity factor is introduced to make the acceptance criterion adaptable. It is calculated using Equation \ref{eqn:sim_factor}, where $f_i$ represents the similarity factor for class $i$, and $s_i$ denotes the similarity of class $i$. This formula is designed so that as similarity increases, the similarity factor decreases. In this formulation, if the similarity score—ranging between 0 and 1—is close to 1, the similarity factor approaches 1 as well, exerting minimal impact on the acceptance level. Conversely, if the similarity score is closer to 0, indicating low similarity, the similarity factor will be higher, raising the acceptance threshold and thus tightening the acceptance criteria. The primary purpose of the similarity factor is to regulate the acceptance mechanism, particularly in cases of low similarity where numerous instances might otherwise qualify as candidates, which is not desirable.

\begin{equation}
\label{eqn:sim_factor}
f_i = (1/s_i)^{0.5}
\end{equation}

The next step is training a classifier to obtain initial performance measures as a baseline to calculate the number of required instances. Then the algorithm will loop for the number of iterations to oversample the dataset. It is another parameter of the algorithm to control run time. The loop starts by calculating the number of required instances to balance the classes to have a balanced dataset. The ideal balance ratio, the ratio of instances in classes, is 1 according to \cite{amin2016comparing} in ideal cases. However, in the real world, it is very unlikely to have such a ratio, especially for some problems where the positive class is a very rare event like fraud detection, disease detection, etc. So, reaching the ideal scenario is almost impossible and the ideal value of the balance ratio is still a big question. We defined the balance ratio as a parameter to find it experimentally. The formula to calculate the number of required instances to balance a class is given in Equation  \ref{eqn:balance_ins}. The output of this formula can be negative for the classes there is no need for oversampling. Our algorithm will ignore these classes and will only focus on the classes that need to be oversampled. In Equation \ref{eqn:balance_ins}, the $n$ represents the total number of instances in the dataset. $r$ is the balance ratio and $n_i$ stands for the number of instances for the minority class. With formula \ref{eqn:balance_ins} it is possible to calculate the number of required instances to balance a dataset numerically.

\begin{equation}
\label{eqn:balance_ins}
(n \cdot r - n_i) \cdot 2
\end{equation}

In some cases, while an imbalance problem in the dataset arises, the performance measures still can be good. Despite the dataset suffering from the imbalance problem, the output of the classification does not suffer. So it is not necessary to oversample that class to be in search of performance improvement. Or, congruently, the required instance numbers can be adjusted by the performance measure. So, to focus only on the classes that have poor performance in the classification task because of the imbalance, we introduced Equation \ref{eqn:balance_ratio} to calculate the required number of instances. The Equation \ref{eqn:balance_ratio}takes into account the performance measure within $\rho$ which affects the number of required instances. The new formula is given in Equation \ref{eqn:balance_ratio} which incorporates the performance factor as $\rho$. If the performance measure increases $\rho$ needs to decrease and vice versa to be able to reflect the effect of the performance measure on the formula. 

\begin{equation}
\label{eqn:balance_ratio}
max(0, (n \cdot r - n_i) \cdot 2 \cdot \rho)
\end{equation}

The performance factor $\rho$ is set to $1 - performance$ $measure$ where the performance measure is maximized and equals to $performance$ $measure$ where it is minimized. If the performance measure is far from its ideal score, then the performance factor should be higher to increase the number of required instances. The maximum function is used to ignore if the number of required instances is negative which is for undersampling which is not a concern for our algorithm. At this point, we need to clarify that the number of required instances is not the number of instances that the oversampling algorithm will reach, rather, it will be used as a priority term to calculate the selection probabilities of the classes. The selection probabilities are calculated by normalizing the number of required instances for all classes within the 0-1 range, hence, the class that has the highest number of required instances will have the highest selection probability to give higher chances to the classes that are highly imbalanced and have poor performance measures. At each iteration, a class is selected probabilistically rather than deterministically and this ensures to find instances that have a higher potential to improve the overall performance. If a single class is chosen for a certain number of iterations, overall performance might be harmed while the focused class is being improved. The main aim is to add an exploitation mechanism to diversify the instance set. 

With the focus on the selected class, a batch of potential instances from the unlabeled set is found from the shuffled unlabeled set by calculating the similarity between the instance and the selected class. The instances have higher similarity scores than the similarity threshold which is calculated by multiplying the class similarity and the similarity factor for that class is added to the candidate list. The class similarity is explained in section \ref{proposedmethod}, and the similarity factor, $f_i$, is calculated with Equation \ref{eqn:sim_factor}.

Potential instances are those that are labeled by the algorithm since they contribute to the performance of the classifier. Since our problem is a multi-label classification problem, the labels for the other classes are determined with the same logic. If the similarity of the instances in the batch exceeds the similarity threshold for the other classes, there are also potential instances for those classes as well. So the potential label vectors for the candidate instances are prepared. A classifier is trained with the combination of labeled data and candidate instances with their potential labels. The chosen measure is calculated for the test set with the trained classifier. If the result gets better compared to the previous result, the candidate instances are added to the labeled set. It can be said that the instances that have high similarity to the existing ones help to improve the overall score and can be added to the labeled set. If the performance is not improved then the instances are not added. 

In the next step, after deciding whether the candidate instances are added or not, the similarity factor is updated accordingly. As emphasized previously, the similarity factor controls the filtering mechanism for finding new potential instances. It is used in the calculation of similarity thresholds directly and has the effect of choosing similar instances. If the potential instances help to improve the performance and are added to the labeled set, the similarity factor is increased as indicated in Equation \ref{eqn:upd_sim_factor1}. Otherwise, if the instances do not help performance improvement and are not added to the labeled set, the similarity factor is decreased to loosen the acceptance filter (Equation \ref{eqn:upd_sim_factor2}).

\begin{equation}
\label{eqn:upd_sim_factor1}
    f_i = f_i \cdot (1 + (1-f_i)^4) 
\end{equation}
\begin{equation}
\label{eqn:upd_sim_factor2}
    f_i = f_i \cdot (1 - (1-f_i)^4)
\end{equation}

The last step in the iteration is updating the iteration number and removing the candidate instances from the unlabeled set. No matter whether the candidate instances are added or not to the labeled set, they are removed from the unlabeled set. So that the algorithm will not make calculations for the same instances every time. It is because the algorithm checks for all potential labels that can be given to them in the previous step and clears off the possibility of being a potential candidate in the next iterations.

\begin{figure}[H]
	\center
    \includegraphics[width=1\textwidth]{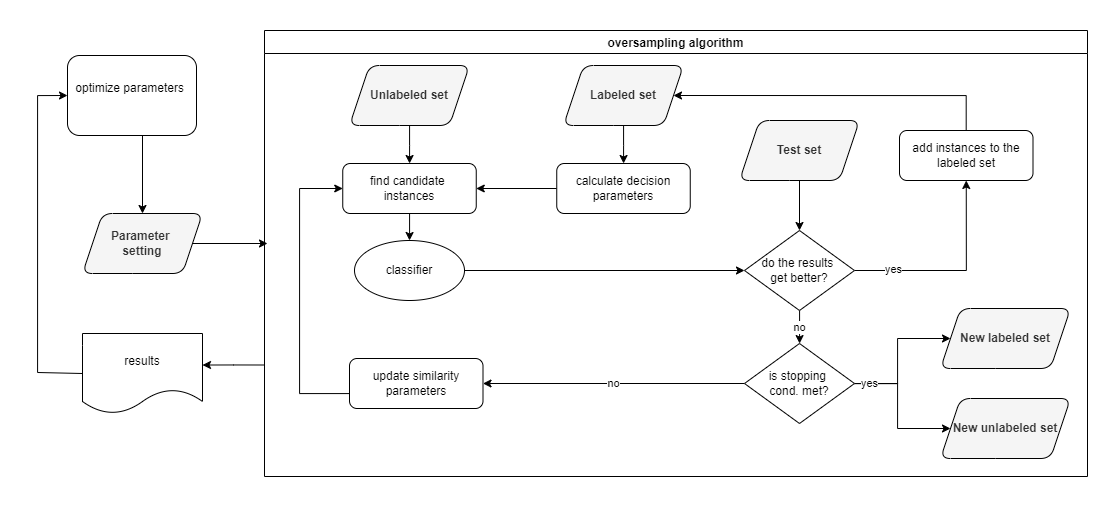}
    \caption{Flow chart for oversampling algorithm.}
    \label{fig:oversampling_alg}
\end{figure}

Once the oversampling algorithm is completed, the oversampled data is returned to the overall algorithm. The final classification is trained with the oversampled data to compare the performance obtained with the initial version of the data. A flow chart of the algorithm can be seen in Figure \ref{fig:oversampling_alg}.

\section{Experimental Results}
\label{chp:b4}

In order to test our proposed algorithm, we conduct experiments and perform analyses using the OPP-115 dataset. The OPP-115 dataset is a collection of website privacy policies to analyze the data practices \cite{wilson2016creation}. Website privacy policies are plain texts formed by a group of paragraphs that cover some topics related to regulations, collection of user data, sharing, processing, data security, etc. Each paragraph is related to one or more topics which converts the problem into a multi-label, multiclass classification problem.

In Table \ref{datastat}, the summary statistics for the dataset are given. Cardinality is defined as the average number of labels per example and density is the average number of labels for each sample obtained by dividing by the total number of labels \cite{bernardini2014cardinality}. Also, the average number of words per instance is given. In the last column, the maximum imbalance ratio is given to demonstrate how imbalanced the dataset is.

\begin{table}[h]
\caption{Summary statistics for the Opp-115 dataset.}\label{datastat}
\begin{tabular*}{\textwidth}{@{\extracolsep{\fill}}lcccccc}
\toprule
Domain & \# of labels & \# of instances & Cardinality & Density & AWC & MIR \\
\midrule
Legal text & 12 & 3,399 & 1.19 & 0.099 & 82 & 38:1 \\ 
\bottomrule
\end{tabular*}
\footnotetext[]{AWC: Average Word Count, MIR: Maximum Imbalance Ratio}
\end{table}

\begin{figure}[h]
	\center
    \includegraphics[width=1\textwidth]{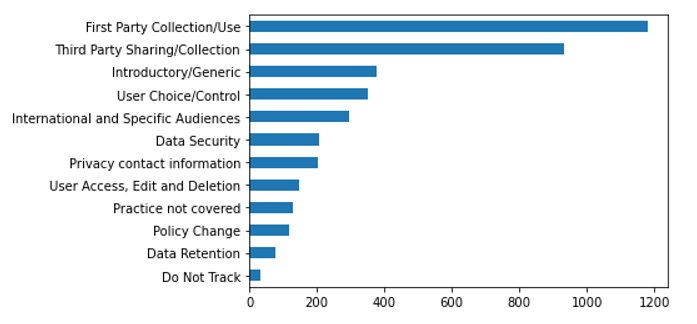}
    \caption{The distribution of the classes.}
    \label{opp115dist}
\end{figure}

The dataset has 12 unique labels and each instance takes at least 1 and at most 5 different labels. The number of instances belonging to each class is not uniformly distributed, while a class has 1,181 instances, others have very few numbers like 31 and 78, which is represented in Figure \ref{opp115dist}. 

Our oversampling algorithm has several parameters that affect the performance of the algorithm. To test the proposed algorithm whether helps to improve performance or not, the dataset is split into labeled and unlabeled sets. The labeled set will be used to train and test the algorithm and the unlabeled set will be used to find new instances. The k-fold cross-validation technique where k equals 5 is employed to validate the method on all the parts of the dataset. 

The response variable is defined as the \textbf{percentage improvement} in the chosen measure after oversampling. It is calculated with the initial measure obtained with the baseline classifier before oversampling and the final measure obtained with the final classifier after oversampling with Equation \ref{eqn:perf_imp}. In this equation, $p_0$ represents the initial performance measure and $p_1$ represents the final performance measure. It is needed to clarify that, both the initial and final performance measures should be included in the response variable. We aim to evaluate the contribution of the algorithm to the performance measure by oversampling the dataset. So, we need to include the initial scores to see the effect.

\begin{equation}
     \%\ improvement = \frac{p_1 - p_0}{p_0} 
    \label{eqn:perf_imp}
\end{equation}

In order to find the best values for the algorithm parameters, a variety of values has been included in the parameter optimization phase. The algorithm has five parameters in total. Two of them are categorical and can take a limited number of values. We tried all possible options for categorical parameters and some values that are chosen intuitively are tried using the grid search technique. As a summary, the final values for the parameters are listed in Table \ref{opt_param_settings}.

\begin{table}[h]
\centering
\renewcommand{\arraystretch}{1.5} 
\setlength\tabcolsep{8pt}         
\caption{Parameter settings for parameter optimization.}
\label{opt_param_settings}
\begin{tabular*}{\textwidth}{@{\extracolsep{\fill}}p{0.5\textwidth} p{0.5\textwidth}@{}}
    \toprule
    \textbf{Parameter} & \textbf{Values} \\ 
    \midrule
    Balance ratio                 & 0.2, 0.3, 0.4, 0.5           \\ 
    Similarity calculation type   & average, safe\_interval       \\ 
    Batch size                    & 1, 2, 3, 5, 7                 \\ 
    Number of iterations          & 50, 100, 200, 500             \\ 
    Similarity type               & euclidean, cosine, jensen-shannon \\ 
    \bottomrule
\end{tabular*}
\end{table}

Using the OPP-115 dataset, we determined the optimal values for the algorithm parameters, and the final results, based on these values, are presented in Table \ref{results_table}. The best performance was achieved with a $batch\ size$ of $5$, a $similarity\ type$ of $euclidean$, a $similarity\ calculation\ type$ of $safe\ interval$, a $balance\ ratio$ of $0.2$, and a $number\ of\ iterations$ set to $100$. The average F1-score before oversampling was $0.5961$, which increased to $0.628$ after applying the oversampling algorithm, representing a 5.34\% improvement. A total of $90$ instances were added, with the initial sample size increasing from $135$ to $225$. The computational results demonstrate that the proposed methodology effectively identifies and labels unlabeled instances that most significantly contribute to classifier performance, thereby enhancing the training set and leading to improved overall performance. The addition of these new instances via the oversampling algorithm resulted in a notable 5.34\% increase in the classifier’s F1-score.

\begin{table}[h]
\centering
\renewcommand{\arraystretch}{1.5} 
\setlength\tabcolsep{8pt}         
\caption{Summary of Results}
\label{results_table}
\begin{tabular}{@{}p{0.5\textwidth} p{0.45\textwidth}@{}}
    \toprule
    \textbf{Metric} & \textbf{Value} \\ 
    \midrule
    Initial F1-score & 0.5961 \\
    Final F1-score after oversampling & 0.6280 \\
    Improvement in F1-score & 5.34\% \\
    Initial number of instances & 135 \\
    Number of instances added by oversampling & 90 \\
    Final number of instances after oversampling & 225 \\
    Best batch size & 5 \\
    Best similarity type & Euclidean \\
    Best similarity calculation type & Safe interval \\
    Best balance ratio & 0.2 \\
    Best number of iterations & 100 \\
    \bottomrule
\end{tabular}
\end{table}

\begin{figure}[h]
	\center
    \includegraphics[width=1\textwidth]{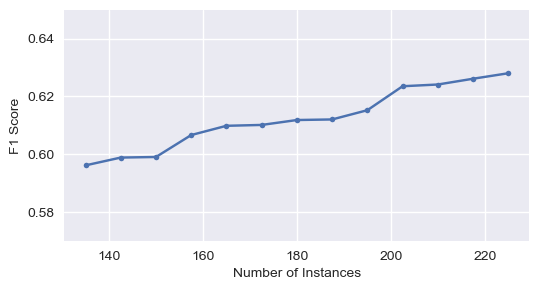}
    \caption{Learning curve: Impact of oversampling on F1 score }
    \label{learningcurve}
\end{figure}

While taking the runs to optimize parameters, we also measure the run times to evaluate the performance of the method. All runs are implemented on a personal computer that has AMD Ryzen 9 3950X CPU and 64 GB memory. The average run time is 2.34 minutes with a standard deviation of 0.71 minutes for 100 iterations. We can say that the algorithm finds new instances from the unlabeled set in a very short period of time. Compared to manual data annotation, a method that finds new instances in minutes can be considered efficient. The learning curve is as in Figure \ref{learningcurve}.

\section{Conclusion and Future Research Directions}
\label{conclusion}
This study presents a similarity-based oversampling algorithm to address the pervasive class imbalance issue in multi-label text classification tasks. The algorithm seeks to expand the labeled dataset by identifying similar instances to the labeled instances within an unlabeled dataset, improving model performance without relying on synthetic data generation. Through a systematic, iterative process, the algorithm identifies instances that exhibit high similarity to the labeled instances, using a similarity measure such as Euclidean distance, cosine similarity, and Jensen-Shannon distance. In our experiments, the algorithm demonstrated good performance in locating high-quality, label-appropriate instances. The results showed that incorporating these instances into the labeled set led to substantial improvements in classifier performance, with a significant positive change in the F1 score. This approach addresses a critical gap in multi-label classification by offering a practical and scalable oversampling solution that leverages real instances, mirroring the benefits of human annotation.

In contrast to many self-learning or semi-supervised methods, which rely only on initial classifier predictions to label new instances, this algorithm considers the risk of inaccurate labeling by inspecting the contribution of new instances on performance. Therefore, it leans on similarity measures to find potential instances, which provides a robust, performance-based foundation for extending the labeled set. By focusing on real instances that align with the existing data distribution, this method helps enrich the dataset while maintaining the integrity of class relationships. Overall, the proposed algorithm contributes a novel perspective to the literature on oversampling in multi-label text classification, serving as a potential baseline for future studies.

There are several avenues for advancing this study. First of all, the choice of similarity function plays a crucial role in the accuracy of instance selection; thus, further research could focus on investigating the effect of similarity metrics on performance. Exploring alternative similarity functions that capture subtle contextual and semantic nuances within text data or designing a new similarity metric specifically tailored for multi-label text data could be one of the open doors to enhancing performance.  Second, it would be valuable to test this method on diverse datasets across multiple domains to better understand the relationship between dataset characteristics and parameter settings. Such research could lead to recommendations on optimal parameter configurations for different types of datasets, thereby enhancing the algorithm’s versatility and adaptability.

Integrating active learning with the similarity-based oversampling algorithm represents another promising direction. In cases where the algorithm has low confidence in assigning labels to certain instances, these instances could be flagged for review by domain experts. This selective labeling strategy would streamline the labeling process, ensuring high-quality labels while reducing the overall effort required by human annotators. Furthermore, the proposed method could be adapted for semi-supervised or self-supervised learning applications, where it would leverage minimal labeled data to support performance improvements in resource-limited environments.

Finally, our study highlights the importance of choosing appropriate performance metrics in multi-label classification, particularly for imbalanced data. While classification metrics such as the F1 score provide valuable insights, ranking-based metrics designed for multi-label settings could offer additional perspectives, especially in scenarios where label ranking is critical. However, these metrics remain underexplored in the context of imbalanced multi-label data due to challenges in interpretability and sensitivity. Future research could investigate the efficacy of ranking-based metrics more closely, potentially developing interpretative frameworks or adaptations that make these metrics more applicable to imbalanced multi-label classification.

\section*{Declarations}

\textbf{Author contributions:} Conceptualization, I.H.K., G.K. and L.E.; methodology, I.H.K., G.K. and L.E.; software, I.H.K.; validation, I.H.K., G.K. and L.E.; formal analysis, I.H.K.; investigation, I.H.K., G.K. and L.E.; data curation, I.H.K. and S.S.; writing---original draft preparation, I.H.K. and S.S.; writing---review and editing, I.H.K., G.K., L.E. and S.S.; visualization, I.H.K. and S.S.; supervision, G.K. and L.E. All authors have read and agreed to the published version of the manuscript. \\

\noindent
\textbf{Funding:} This research received no external funding. \\

\noindent
\textbf{Data availability:} The dataset used in this work is publicly available. \\

\noindent
\textbf{Conflict of interest:} The authors declare no conflict of interest or competing interest. \\

\noindent
\textbf{Ethics approval and consent to participate:} Not applicable.

\begin{appendices}

\section{}\label{app1}

\textbf{Huggingface Embeddings}

\begin{itemize}
    \item 'stsb-roberta-large'
    \item 'all-MiniLM-L6-v2'
    \item 'all-MiniLM-L12-v2'
    \item 'all-mpnet-base-v1'
    \item 'all-mpnet-base-v2'
    \item 'all-roberta-large-v1'
    \item 'all-distilroberta-v1'
    \item 'albert-base-v2'
    \item 'ALBERT-xxlarge'
    \item 'bert-base-nli-mean-tokens'
    \item 'all-roberta-large-v1'
    \item 'distiluse-base-multilingual-cased-v1'
    \item 'multi-qa-mpnet-base-dot-v1'
    \item 'all-distilroberta-v1'
    \item 'bert-base-uncased'
    \item 'bert-base-nli-mean-tokens'
    \item 'distiluse-base-multilingual-cased-v1'
    \item 'distilbert-base-nli-mean-tokens'
    \item 'multi-qa-mpnet-base-dot-v1'
    \item 'nlpaueb/legal-bert-base-uncased'
    \item 'paraphrase-multilingual-MiniLM-L12-v2'
    \item 'paraphrase-mpnet-base-v2'
    \item 'paraphrase-MiniLM-L6-v2'
    \item 'paraphrase-xlm-r-multilingual-v1'
    \item 'saibo/legal-roberta-base'
    \item 'ALBERT-xlarge'
    \item 'sentence-t5-large'
    \item 'sentence-transformers/average-word-embeddings-glove.6B.300d'
    \item 'sentence-transformers/average-word-embeddings-glove.840B.300d'
\end{itemize}

\textbf{OpenAI Embeddings}

\begin{itemize}
    \item 'text-similarity-babbage-001'
    \item 'text-similarity-ada-001'
    \item 'text-similarity-curie-001'
    \item 'text-similarity-davinci-001'
\end{itemize}

\textbf{Google Embeddings}

\begin{itemize}
    \item 'universal-sentence-encoder'
\end{itemize}

\end{appendices}

\bibliography{sn-bibliography}

\end{document}